\documentclass{bmvc2k}
\usepackage{wrapfig}
\usepackage{graphicx}
\usepackage[b5,center]{crop}
\makeatletter
\renewcommand{\BMVA@blfootnote}[1]{}  % makes it do nothing
\makeatother
\usepackage{enumitem}
\usepackage{booktabs}
\usepackage{multicol}
\usepackage{multirow}
\title{\raisebox{-0.35em}{\includegraphics[height=1.5em]{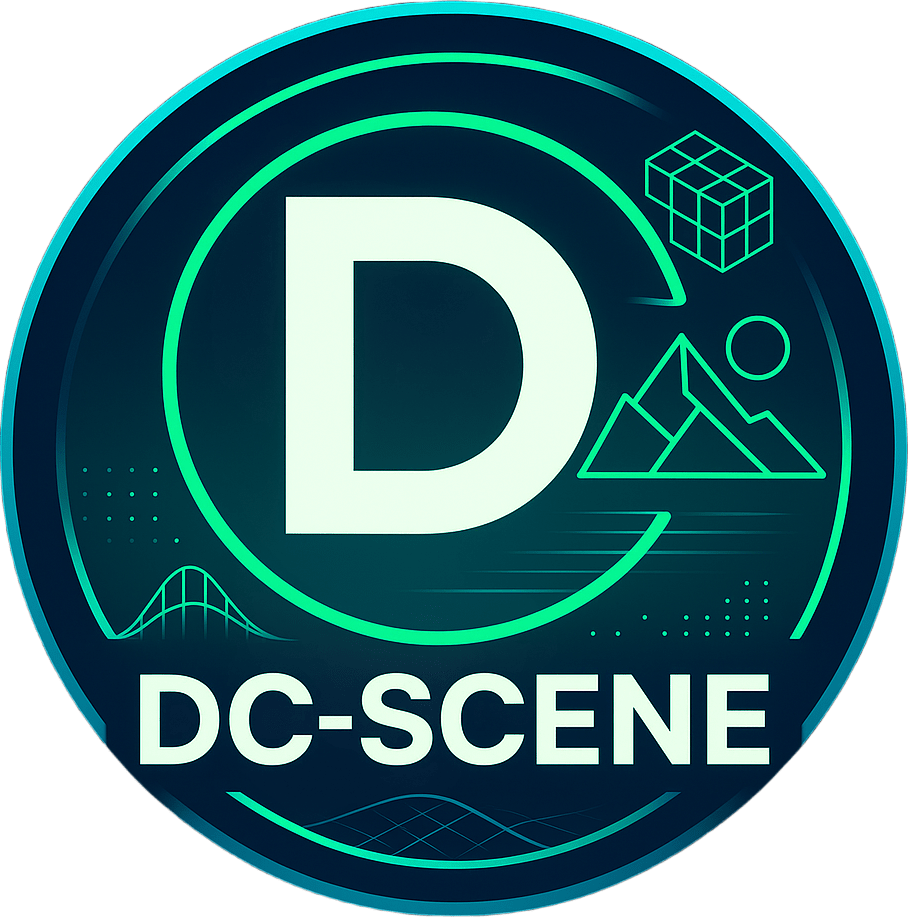}}~DC-Scene: Data-Centric Learning for 3D Scene Understanding}

\addauthor{Ting Huang$^*$}{huangting0403@sues.edu.cn}{1}
\addauthor{Zeyu Zhang$^{*\dag}$}{https://steve-zeyu-zhang.github.io}{2}
\addauthor{Ruicheng Zhang}{zhangrch23@mail2.sysu.edu.cn}{3}
\addauthor{Yang Zhao$^\ddag$}{y.zhao2@latrobe.edu.au}{4}

\addinstitution{
 Shanghai University of Engineering\\Science
}
\addinstitution{
 The Australian National University
}
\addinstitution{
 Sun Yat-sen University
}
\addinstitution{
 La Trobe University\\
 ~\\
 \small $^*$Equal contribution. $^\dag$Project lead.\\
 \small $^\ddag$Corresponding author.
}

\runninghead{Ting Huang and Zeyu Zhang et al.}{DC-Scene}

\begin{document}
% 9 page + reference

\maketitle

\begin{abstract}
3D scene understanding plays a fundamental role in vision applications such as robotics, autonomous driving, and augmented reality. However, advancing learning-based 3D scene understanding remains challenging due to two key limitations: (1) the large scale and complexity of 3D scenes lead to higher computational costs and slower training compared to 2D counterparts; and (2) high-quality annotated 3D datasets are significantly scarcer than those available for 2D vision. These challenges underscore the need for more efficient learning paradigms. In this work, we propose \textbf{DC-Scene}, a data-centric framework tailored for 3D scene understanding, which emphasizes enhancing data quality and training efficiency. Specifically, we introduce a CLIP-driven dual-indicator quality (DIQ) filter, combining vision-language alignment scores with caption-loss perplexity, along with a curriculum scheduler that progressively expands the training pool from the top 25\% to 75\% of scene–caption pairs. This strategy filters out noisy samples and significantly reduces dependence on large-scale labeled 3D data. Extensive experiments on ScanRefer and Nr3D demonstrate that DC-Scene achieves state-of-the-art performance (\textbf{86.1 CIDEr with the top-75\% subset vs. 85.4 with the full dataset}) while reducing training cost by approximately two-thirds, confirming that a compact set of high-quality samples can outperform exhaustive training.
Code will be available at \url{https://github.com/AIGeeksGroup/DC-Scene}.
\end{abstract}
\vspace{-0.5cm}

%-------------------------------------------------------------------------
\section{Introduction}
\label{sec:intro}

% \begin{figure}[t]
%     \centering
%     \includegraphics[width=\linewidth]{images/main.pdf}
%     \caption{Effect of CLIP-score–based data filtering on training cost. Horizontal bars report the cumulative GPU-hours required to fine-tune 3D CoCa when retaining only the top-$k$ percent of scene–caption pairs ranked by CLIP similarity. Filtering out the lowest-quality 80\% of samples cuts training time by 75\%, demonstrating the efficiency gains that motivate our data-centric learning strategy.}
%     \label{fig:main}
%     \vspace{-0.4cm}
% \end{figure}

\begin{wrapfigure}{r}{0.5\linewidth}
    \centering
    % \vspace{-0.7cm}
    \vspace{-1.5cm}
    \includegraphics[width=\linewidth]{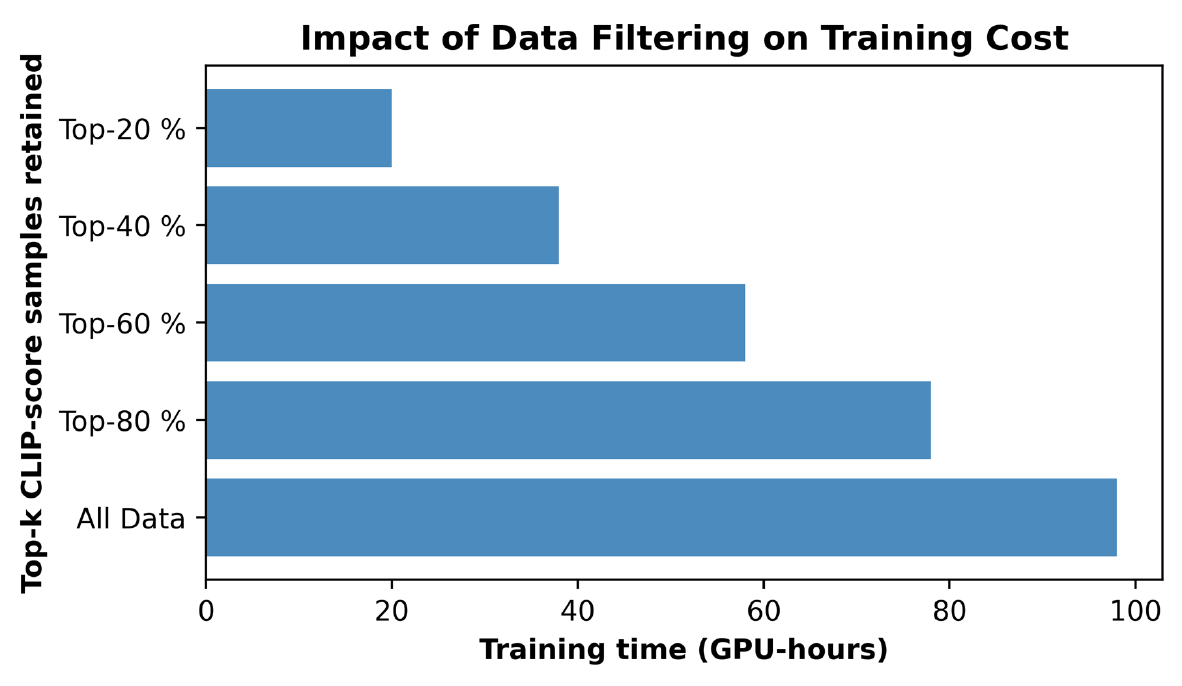}
    \caption{\textbf{Effect of CLIP-score–based data filtering on training cost.} Filtering out the lowest-quality 80\% of samples cuts training time by 75\%, motivating our data-centric learning strategy.}
    \label{fig:main}
    \vspace{-1.3cm}
\end{wrapfigure}

3D scene understanding plays a vital role in a wide range of real-world applications, including robotics, augmented reality(AR), virtual reality(VR), and autonomous driving~\cite{wang2024open,sportscap2021,chen2021tightcap,Liao_Zhu_Zhang_Ye_Chen_Fan_2021,dcnet2023}. For instance, robots must interpret complex 3D environments to navigate and interact safely, while AR systems rely on accurate scene interpretation to seamlessly integrate virtual content with the physical world. Achieving detailed 3D scene understanding entails recognizing objects, analyzing their spatial relationships, and generating natural language descriptions—a task known as 3D scene captioning. High-quality 3D scene captioning enables robots to perform sophisticated tasks and empowers AR systems to deliver context-aware user experiences.

However, despite recent advances in 3D scene understanding~\cite{todocap2025,huang20253d,li2024dense,Fu2025SceneLLM,jia2024sceneverse}, significant challenges remain that hinder the efficiency and effectiveness of current methods. The first major challenge stems from the high computational cost associated with training on 3D data. Unlike conventional 2D images, 3D representations such as dense point clouds or volumetric meshes are substantially larger and more complex, resulting in considerably greater computational demands and slower training times, as illustrated in Figure~\ref{fig:main}. A second prominent limitation is the scarcity of high-quality annotated 3D datasets. While the availability of large-scale labeled datasets has driven progress in 2D vision, annotated 3D data remains limited due to the labor-intensive process of capturing and labeling complex scenes. For instance, ScanRefer~\cite{chen2020scanrefer}, a widely used dataset for 3D captioning, contains annotations for only around 800 scenes which is far fewer than typical 2D datasets. This data scarcity often leads to model overfitting and poor generalization, particularly in complex or real-world environments.

These challenges have motivated a shift toward data-centric learning methods, which prioritize the quality and structure of training data over the development of increasingly complex neural architectures. To address the computational inefficiencies inherent in processing large-scale 3D data, we propose a data-centric approach that optimizes training efficiency and effectiveness. Specifically, we leverage pretrained vision-language embeddings from CLIP to systematically evaluate the quality of each 3D scene–caption pair. By prioritizing semantically aligned, high-quality samples, our method substantially accelerates convergence and reduces computational overhead. In parallel, to address the scarcity of labeled 3D data, we introduce a dynamic curriculum learning strategy that gradually incorporates more challenging or noisy samples into the training process. Starting from clearer and simpler examples, this staged curriculum mitigates overfitting and improves generalization to diverse and complex 3D scenes encountered in real-world applications.

We integrate these data-centric strategies into a unified framework termed \textbf{DC-Scene}. While we demonstrate the effectiveness of DC-Scene using the state-of-the-art 3D CoCa architecture~\cite{huang20253d}, the proposed methods are not limited to this specific backbone. Instead, they constitute a flexible and generalizable paradigm that enhances training efficiency and performance across a range of foundational architectures. In doing so, DC-Scene offers a robust and adaptable solution to the dual challenges of computational overhead and limited data availability in 3D scene understanding, thereby contributing broadly to the advancement of the field.

In summary, our contributions are as follows:
\begin{itemize}[topsep=0pt,partopsep=0pt,itemsep=0pt,parsep=0pt]
    \item We introduce a novel CLIP-driven module for scoring data quality, effectively addressing the critical issues of variability and noise in 3D scene–caption pairs.
    \item We propose \textbf{DC-Scene}, a flexible and data-centric curriculum learning framework that enhances training efficiency and generalizability, and can be seamlessly integrated into various 3D scene captioning backbones.
    \item We conduct extensive experiments demonstrating accelerated convergence and state-of-the-art captioning performance on ScanRefer~\cite{chen2020scanrefer} (86.10 CIDEr@0.25) and Nr3D~\cite{achlioptas2020referit_3d} (53.60 CIDEr@0.50), requiring only one-third of the training epochs compared to full-data baselines.
\end{itemize}

\section{Related Works}
\paragraph{3D Scene Understanding}
3D scene understanding primarily encompasses tasks such as 3D visual question answering (VQA)~\cite{ma2022sqa3d,azuma2022scanqa} and 3D dense captioning~\cite{chen2020scanrefer,achlioptas2020referit_3d}. Although this work focuses on 3D scene captioning, the proposed techniques have the potential to be extended to 3D VQA. Early approaches to 3D scene captioning typically follow a detect-then-describe pipeline, wherein a 3D object detector first localizes objects within the scene, followed by a language model that generates a caption for each detected object~\cite{scan2cap_2021,MORE_2022,spa2cap2022}. For instance, Scan2Cap~\cite{scan2cap_2021} pioneered the task of dense captioning in 3D scenes by detecting objects~\cite{zhao2025peddet,cai2024msdet,cai2024medical,zhang2024meddet} and generating corresponding textual descriptions. Subsequent works improved upon this two-stage framework by modeling inter-object relationships: MORE~\cite{MORE_2022} introduced graph-based relational reasoning over detected objects to produce context-aware captions, while SpaCap3D~\cite{spa2cap2022} employed a spatially guided transformer to better capture spatial relations among object proposals. Despite their effectiveness, these methods rely heavily on precomputed 3D proposals and separate captioning modules, which may propagate detection errors to the caption generation stage.

Recent work has shifted toward end-to-end and transformer-based architectures for 3D captioning~\cite{lee2025duoduo,vote2cap2023,vote2cap++2024,chen2021d3net}. Vote2Cap-DETR~\cite{vote2cap2023} exemplifies this trend as a one-stage, fully transformer-based model that eliminates the need for explicit proposal generation. It formulates 3D captioning as a direct set prediction problem, where a transformer decoder produces a set of queries that are simultaneously processed by a localization head and a caption generation head. Another notable approach, D3Net~\cite{chen2021d3net}, introduces a unified speaker–listener framework to jointly address 3D captioning and 3D visual grounding. By sharing representations across both tasks, D3Net enables semi-supervised training and promotes the generation of more distinctive object descriptions.

The current state-of-the-art in 3D scene captioning is 3D CoCa~\cite{huang20253d}, which introduces a single-stage model that jointly optimizes a contrastive vision-language objective and a captioning objective within a unified framework. It employs a frozen CLIP model as the backbone to inject rich semantic priors; in particular, a CLIP-based vision-language encoder is used to align 3D scene features with corresponding textual representations. By integrating contrastive learning with direct caption generation, 3D CoCa achieves enhanced cross-modal alignment without relying on external object proposals. The success of 3D CoCa underscores the advantages of leveraging pretrained vision-language models to mitigate data sparsity, as well as the benefits of end-to-end training for achieving both spatial and semantic coherence.

\paragraph{Data Centric Learning}
Curriculum learning has emerged as a powerful paradigm in machine learning, enabling models to progress from simpler to more complex tasks in a structured manner. Inspired by the principles of human learning, this approach has been widely applied across domains such as natural language processing and computer vision~\cite{bengio2009curriculum,soviany2022curriculum}. Bengio et al.~\cite{bengio2009curriculum} pioneered the idea that organizing training data in a meaningful progression can significantly improve the learning dynamics and performance of neural networks. Building upon this foundation, Soviany et al.~\cite{soviany2022curriculum} proposed an adaptive curriculum learning strategy that dynamically adjusts the complexity of training samples based on the model’s real-time performance.

Recent studies have extended curriculum learning to 3D vision tasks, aiming to improve model training by sequencing data from easy to hard. For instance, Curricular Object Manipulation (COM)~\cite{zhu2023curricular} incorporates a curriculum strategy into LiDAR-based 3D object detection by progressively emphasizing loss and data augmentation on increasingly difficult objects throughout the training process. Other data-centric methods leverage self-paced or uncertainty-guided sample selection to refine training data and mitigate noise. In semi-supervised 3D object detection, SPSL-3D~\cite{an2023leveraging} employs a self-paced learning framework to weigh and filter pseudo-labeled samples based on their reliability, enabling the model to learn from high-confidence examples while postponing the incorporation of noisy ones. Moreover, sample filtering and active selection techniques have shown promise in enhancing 3D captioning and segmentation. DiffuRank~\cite{luo2024view}, for example, utilizes a pretrained text-to-3D diffusion model to rank rendered views of a 3D object by their alignment with textual descriptions. Captions are then generated using only the top-ranked views, significantly reducing the occurrence of hallucinated content.

\section{Methodology}
\begin{figure}[t]
    \centering
    \vspace{-0.1cm}
    \includegraphics[width=\linewidth]{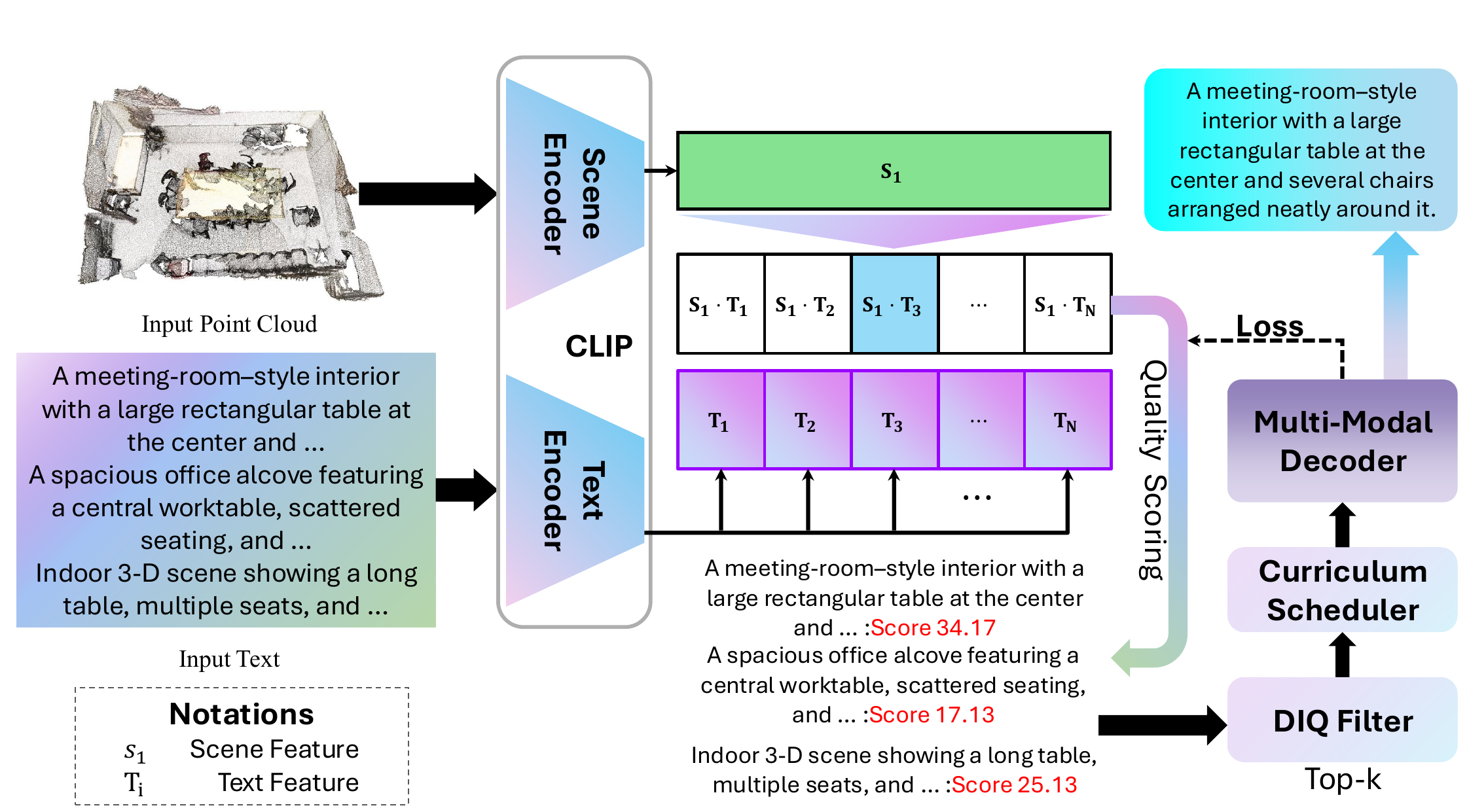}
    \vspace{-0.7cm}
    \caption{\textbf{Framework of DC-Scene.} Raw point cloud scene and candidate captions are first encoded by a Scene Encoder and a Text Encoder. Quality Scoring module computes the CLIP score for each scene–text pair. Dual-Indicator Quality (DIQ) Filter then selects samples that lie within a predefined quality region, retaining only the top-$k$ candidates per scene. These filtered representations are passed to the Curriculum Scheduler, which gradually feeds them into the Multi-Modal Decoder for caption generation. A feedback loop returns the caption loss and updated CLIP scores to refine the quality map, thereby closing the data-centric learning cycle.}
    \label{fig:pipeline}
    \vspace{-0.9cm}
\end{figure}
\subsection{Overview}
In this section, we introduce \textbf{DC-Scene}, a data-centric framework designed to optimize 3D scene captioning through a curriculum learning strategy. As illustrated in Figure~\ref{fig:pipeline}, the framework comprises three key components: (1) quality scoring; (2) a dual-indicator quality (DIQ) filter; and (3) a curriculum scheduler.

\subsection{Quality Scoring}
To address the inherent variability and noise in 3D scene–caption pairs, we employ CLIP-based embeddings to systematically assess data quality. Specifically, given a set of 3D scene-caption pairs \( D=\left \{ \big ( x_j^s,x_j^t \big )  \right \}_{j=1}^N \), we utilize the CLIP 3D scene encoder, defined as $\mathrm{S}_{clip}\left ( \cdot  \right ) $, to extract a feature vector from the input 3D scene $x_j^s$, and the CLIP text encoder, defined as $\mathrm{T}_{clip}\left ( \cdot  \right )$, to obtain a feature vector from the corresponding caption $x_j^t$. We then compute the dot product of both features to generate a clip score, which we define as $s_j$:
\begin{equation}
    s_j=\mathrm{S}_{clip}\left ( x_j^s \right ) \cdot \mathrm{T}_{clip}\left ( x_j^t \right ) .
\end{equation}
We partition the dataset by first identifying the upper bound $S_{max}$ and lower bound $S_{min}$ of the CLIP scores $s_j$. Specifically, $S_{max}$ and $S_{min}$ are determined empirically based on statistical analysis of the dataset, typically by selecting the 95th percentile as the upper bound and the 5th percentile as the lower bound of the CLIP score distribution. This approach ensures a robust and representative selection of high-quality, semantically aligned scene-caption pairs. Based on these bounds, we select the samples $d_j$ whose scores fall within this range to form a subset, which we refer to as the Data of Intermediate Similarity (DIS):
\begin{equation}
    DIS=\left \{ d_j\vert S_{min}\le s_j \le S_{max} \right \}.
\end{equation}
The CLIP score effectively measures the semantic coherence between a caption and its corresponding 3D scene representation, enabling the identification and selection of high-quality data where the scene and caption are closely aligned.

\subsection{Dual-Indicator Quality (DIQ) Filter}
Inspired by recent advancements in multimodal data selection, we introduce a dual-indicator approach that evaluates each data sample based on two criteria: the previously defined CLIP score and the model-generated caption loss, which also serves as a proxy for perplexity. This loss reflects the discrepancy between the ground-truth caption and the model’s current language modeling preferences. Perplexity captures the complexity and uncertainty associated with generating a caption, and we denote it as $l_j$. Formally, the caption perplexity is computed as:
\begin{equation}
    l_j=-\sum_{t=1}^{T} \log P_{\theta }\left ( Y_{j,t}\vert Y_{j,<t},X_j \right )  ,
\end{equation}
where $X_j$ is the latent encoded features provided by the scene encoder, $Y_j$ is the caption generated by the decoder, and $P_{\theta}$ represents the conditional probability from our captioning decoder.

We partition the dataset by identifying the upper bound $L_{max}$ and lower bound $L_{min}$ of the caption loss values. Specifically, $L_{max}$ and $L_{min}$ are determined empirically based on statistical analysis, such as selecting the 95th percentile for upper bound and the 5th percentile for lower bound of the caption loss distribution, thereby effectively excluding extreme outliers. Using these bounds, we select the samples $d_j$ whose loss falls within this range to construct a subset, which we refer to as the Data of Intermediate Loss (DIL):
\begin{equation}
    DIL=\left \{ d_j\vert L_{min}\le l_j\le L_{max} \right \} .
\end{equation}

Once each data sample is associated with both a CLIP score and a caption loss, we construct a two-dimensional quality space defined by these two attributes. Within this space, we apply upper and lower bounds on both dimensions to select a subset of high-quality samples, which we refer to as the Dual-Indicator Quality (DIQ) region:
\begin{equation}
    DIQ=\left \{ d_j\vert L_{min}\le l_j\le L_{max},S_{min}\le s_j\le S_{max} \right \} .
\end{equation}
The $DIQ$ region encompasses samples that exhibit both strong semantic alignment and moderate linguistic complexity, thereby ensuring a diverse yet effective training set.

\subsection{Curriculum Scheduler}
We propose a data curriculum framework that initiates training with simpler tasks and progressively transitions to more complex ones. Based on the defined $DIQ$ region, we partition the space into uniform blocks using step sizes $\Delta S$ and $\Delta L$, which correspond to the CLIP score and caption loss, respectively. By applying targeted data selection strategies, we control the quality of training samples by incrementally increasing both the CLIP score and loss thresholds. This allows us to structure the learning process into $k$ curriculum stages, where varying quantities of training samples are introduced at each stage to reflect increasing data complexity:
\begin{equation}
    C_k=\left \{ d_j\vert L_p\le l_j,S_p\le s_j \right \} ,
\end{equation}
where
\begin{equation}
    L_p=L_{min}+k\Delta L \text{ and } S_p=S_{min}+k\Delta S.
\end{equation}
As $k$ increases, the curriculum is segmented into multiple phases: Initialization, Intermediate, and Advanced.
\begin{itemize}[left=0pt,topsep=0pt,partopsep=0pt,itemsep=1pt,parsep=0pt]
    \item Initialization Phase (k=0): The model starts with a distribution of high-quality data, focusing on underlying patterns without being overwhelmed by complexity.
    \item Intermediate Phase (k=1): Data quality is improved by increasing the thresholds for clip score and loss, narrowing the candidate region of high-quality data.
    \item Advanced Phase (k=2): The model is exposed to the most challenging data, characterized by higher clip score and model loss, testing its ability to handle complex and less consistent relationships.
\end{itemize}

This step-by-step progression ensures a stable learning trajectory, enhances overall model comprehension, reduces the risk of overfitting to specific data instances, and improves the model’s robustness and generalization capability.

\section{Experiments}

\subsection{Dataset and Evaluation Metrics}
\subsubsection{Datasets}
To evaluate the performance of our 3D scene captioning framework, we conduct experiments on two widely used benchmarks: ScanRefer~\cite{chen2020scanrefer} and Nr3D~\cite{achlioptas2020referit_3d}. ScanRefer comprises 36,665 natural language descriptions covering 7,875 unique objects across 562 scenes. In comparison, Nr3D provides 32,919 descriptions corresponding to 4,664 objects across 511 scenes. Both datasets are built upon the ScanNet~\cite{dai2017scannet} repository, which contains 1,201 RGB-D reconstructed indoor scenes. For evaluation, we utilize 9,508 descriptions referencing 2,068 objects from 141 scenes in ScanRefer, and 8,584 descriptions referring to 1,214 objects from 130 scenes in Nr3D. Notably, all evaluation scenes are drawn from the 312 scenes that constitute the ScanNet validation split.

\subsubsection{Evaluation Metrics}
We evaluate captioning performance using four standard metrics: CIDEr\cite{cider2015}, BLEU-4\cite{bleu2002}, METEOR\cite{meteor2005}, and ROUGE-L\cite{rouge2004}, denoted as C, B-4, M, and R, respectively.

\subsection{Implementation Details}
We adopt 3D CoCa~\cite{huang20253d} and Vote2Cap-DETR++~\cite{vote2cap++2024} as our backbone models, pretraining both on the ScanRefer~\cite{chen2020scanrefer} and Nr3D~\cite{achlioptas2020referit_3d} datasets. Our data-centric approach modifies only the training strategy, leaving the model architectures unchanged. We use the AdamW optimizer with a batch size of 8 and an initial learning rate of $1 \times 10^{-4}$. The curriculum learning process is staged across three phases, with transitions occurring at the 360th and 720th epochs, and training concluding at epoch 1,080. Specifically, during epochs 1–360, the model is trained on the top 25\% of samples selected by the Dual-Indicator Quality (DIQ) filter. From epochs 361–720, the data pool expands to include the top 50\%, and from epochs 721–1,080, the full DIQ-qualified dataset is used. All experiments are conducted on a single NVIDIA RTX 4090 GPU.
\begin{figure}[htbp]
    \centering
    \vspace{-0.3cm}
    \includegraphics[width=\linewidth]{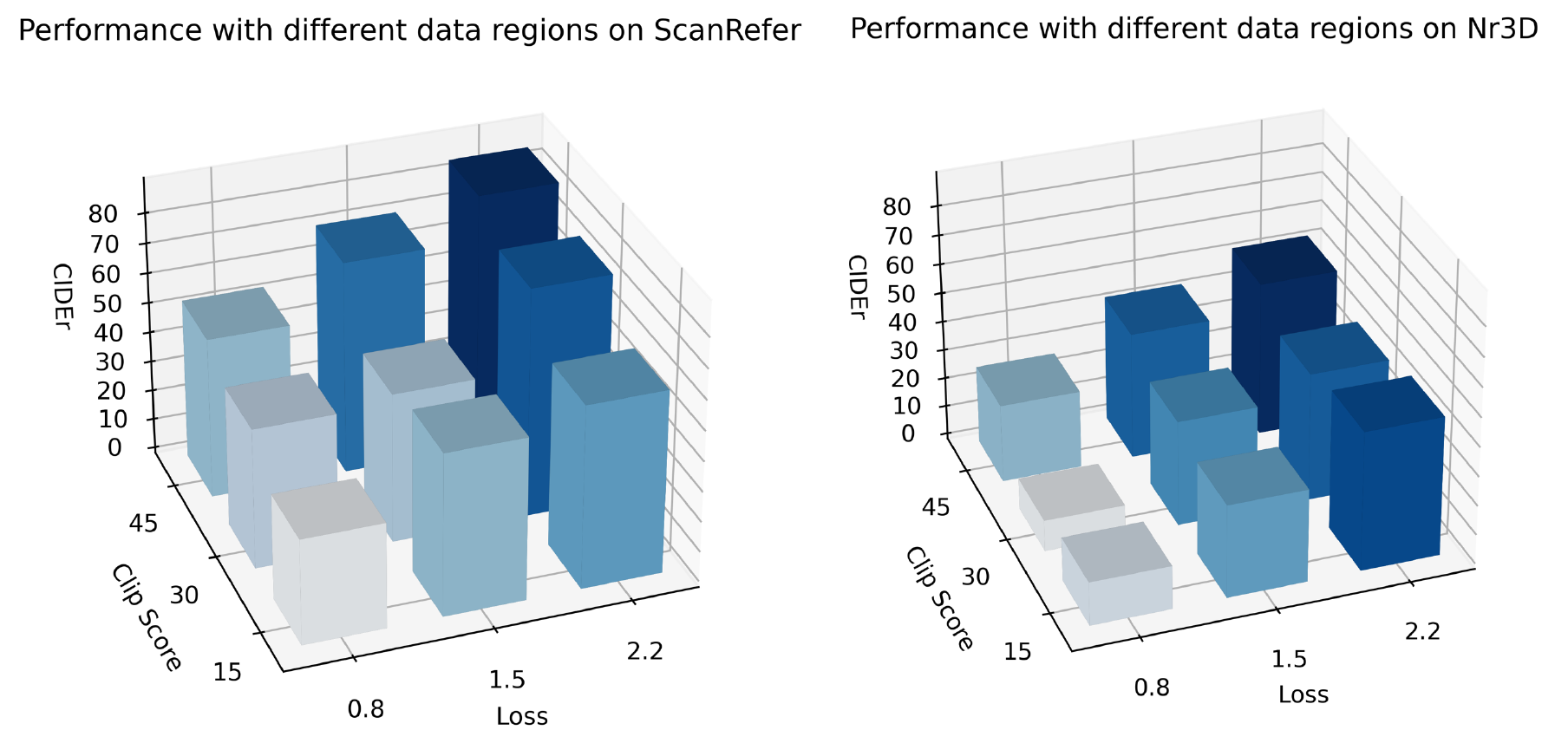}
    \caption{Ablation study comparing different Dual-Indicator Quality (DIQ) regions on the ScanRefer and Nr3D datasets.}
    \label{ablation:diq}
    \vspace{-0.6cm}
\end{figure}
\subsection{Comparative Study}

    \begin{table}[t]
      \centering
      \caption{Results on ScanRefer~\cite{chen2020scanrefer} at IoU = 0.25 and Nr3D~\cite{achlioptas2020referit_3d} at IoU = 0.50, both without additional 2D input. C, B-4, M, and R denote CIDEr~\cite{cider2015}, BLEU-4~\cite{bleu2002}, METEOR~\cite{meteor2005}, and ROUGE-L~\cite{rouge2004}, respectively.}
      \label{tab:comparative}
      \resizebox{\linewidth}{!}{%
      \begin{tabular}{lcccccc}
        \toprule
        Backbone & DIQ ratio & C$\uparrow$ & B-4$\uparrow$ & M$\uparrow$ & R$\uparrow$ & Epochs \\
        \midrule
        \multicolumn{7}{c}{\textbf{ScanRefer}~\cite{chen2020scanrefer}} \\
        \midrule
        \multirow{4}{*}{Vote2Cap-DETR++}
            & 25\,\%               & 70.10 & 39.05 & 27.50 & 58.10 & 360 \\
            & 50\,\%               & 73.24 & \textbf{41.80} & 28.60 & 60.30 & 360 \\
            & 75\,\%               & \textbf{78.55} & 41.20 & \textbf{29.10} & \textbf{60.95} & 360 \\
            & 100\,\% (baseline)   & 76.36 & 41.37 & 28.70 & 60.00 & 1\,080 \\
        \midrule
        \multirow{4}{*}{3D CoCa}
            & 25\,\%               & 82.15 & 44.80 & 30.60 & 60.90 & 360 \\
            & 50\,\%               & 85.60 & 45.80 & 30.89 & 61.80 & 360 \\
            & 75\,\%               & \textbf{86.10} & \textbf{46.00} & \textbf{31.40} & \textbf{62.20} & 360 \\
            & 100\,\% (baseline)   & 85.42 & 45.56 & 30.95 & 61.98 & 1\,080 \\
    \midrule
        \multicolumn{7}{c}{\textbf{Nr3D}~\cite{achlioptas2020referit_3d}} \\
        \midrule
    \multirow{4}{*}{Vote2Cap-DETR++}
        & 25\,\%               & 43.90 & 26.50 & 25.00 & 54.00 & 360 \\
        & 50\,\%               & 47.25 & 27.90 & 25.60 & 55.40 & 360 \\
        & 75\,\%               & \textbf{48.10} & \textbf{28.20} & \textbf{25.70} & \textbf{55.60} & 360 \\
        & 100\,\% (baseline)   & 47.08 & 27.70 & 25.44 & 55.22 & 1\,080 \\
    \midrule
    \multirow{4}{*}{3D CoCa}
        & 25\,\%               & 50.20 & 28.00 & 25.00 & 55.50 & 360 \\
        & 50\,\%               & 53.00 & 29.05 & 25.90 & 56.10 & 360 \\
        & 75\,\%               & \textbf{53.60} & \textbf{29.40} & \textbf{26.00} & \textbf{56.50} & 360 \\
        & 100\,\% (baseline)   & 52.84 & 29.29 & 25.55 & 56.43 & 1\,080 \\
    \bottomrule
  \end{tabular}}
  \vspace{-0.7cm}
\end{table}
We evaluate \textbf{DC‐Scene} using the backbone models 3D CoCa~\cite{huang20253d} and Vote2Cap‐DETR++~\cite{vote2cap++2024}, on the ScanRefer~\cite{chen2020scanrefer} and Nr3D~\cite{achlioptas2020referit_3d} benchmarks.

As shown in Table~\ref{tab:comparative}, the top-75\% DIQ split achieves the best trade-off between training efficiency and captioning accuracy. With only 360 training epochs, 3D CoCa reaches 86.10 CIDEr on ScanRefer and 53.60 CIDEr on Nr3D. A similar trend is observed for Vote2Cap-DETR++. These results confirm that training on a moderate subset of high-quality data can reduce training time by two-thirds while maintaining, and in some cases even improving, overall captioning performance.
\subsection{Ablation Study}
\subsubsection{Effectiveness of DIQ}
To assess the effectiveness of the Dual-Indicator Quality (DIQ) criterion, we analyzed CLIP scores and caption loss values across the entire ScanRefer and Nr3D datasets. Each dataset was partitioned into nine distinct regions based on combinations of score and loss thresholds, and we sampled 7,000 scene–caption pairs from each region using DIQ-based selection.
As illustrated in Figure~\ref{ablation:diq}, the horizontal and vertical axes correspond to the CLIP score and loss ranges, respectively, while each column represents a specific region in this dual-indicator space. The color intensity of each column reflects captioning performance, with darker hues indicating stronger results. Aggregating findings from both datasets, we observe that regions characterized by higher CLIP scores and moderate-to-high loss values yield the best captioning outcomes. This indicates that the upper-right quadrant of the DIQ space contains the highest-quality training samples for 3D scene captioning.

\subsubsection{Effectiveness of quality-driven sampling}
\begin{figure}[htbp]
    \centering
    \vspace{-0cm}
    \includegraphics[width=\linewidth]{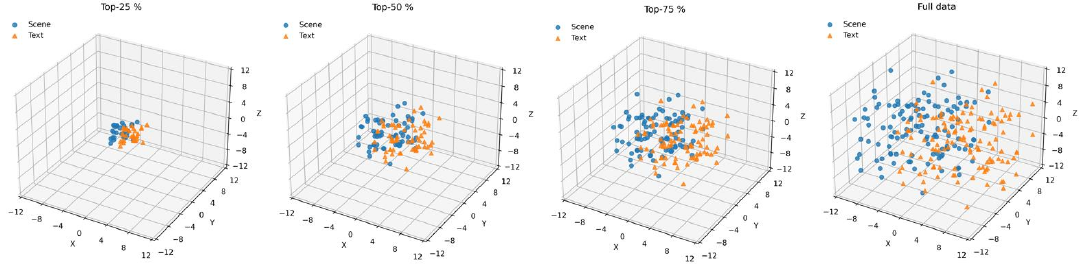}
    \caption{Data quality buckets visualized in 3-D embedding space.}
    \label{ablation:qds}
    \vspace{-0.3cm}
\end{figure}
Figure~\ref{ablation:qds} visualizes the distribution of candidate scene–text pairs used throughout the curriculum in the ScanRefer dataset. The figure illustrates that our CLIP-score filter effectively establishes a natural difficulty continuum—beginning with compact, semantically well-aligned pairs and progressively expanding toward noisier, more challenging samples. Training the model along this structured trajectory facilitates faster convergence and improved generalization, as evidenced by the results in Table~\ref{tab:comparative}.

\subsection{Qualitative Result}
\begin{figure}[htbp]
    \centering
    \vspace{-0cm}
    \includegraphics[width=\linewidth]{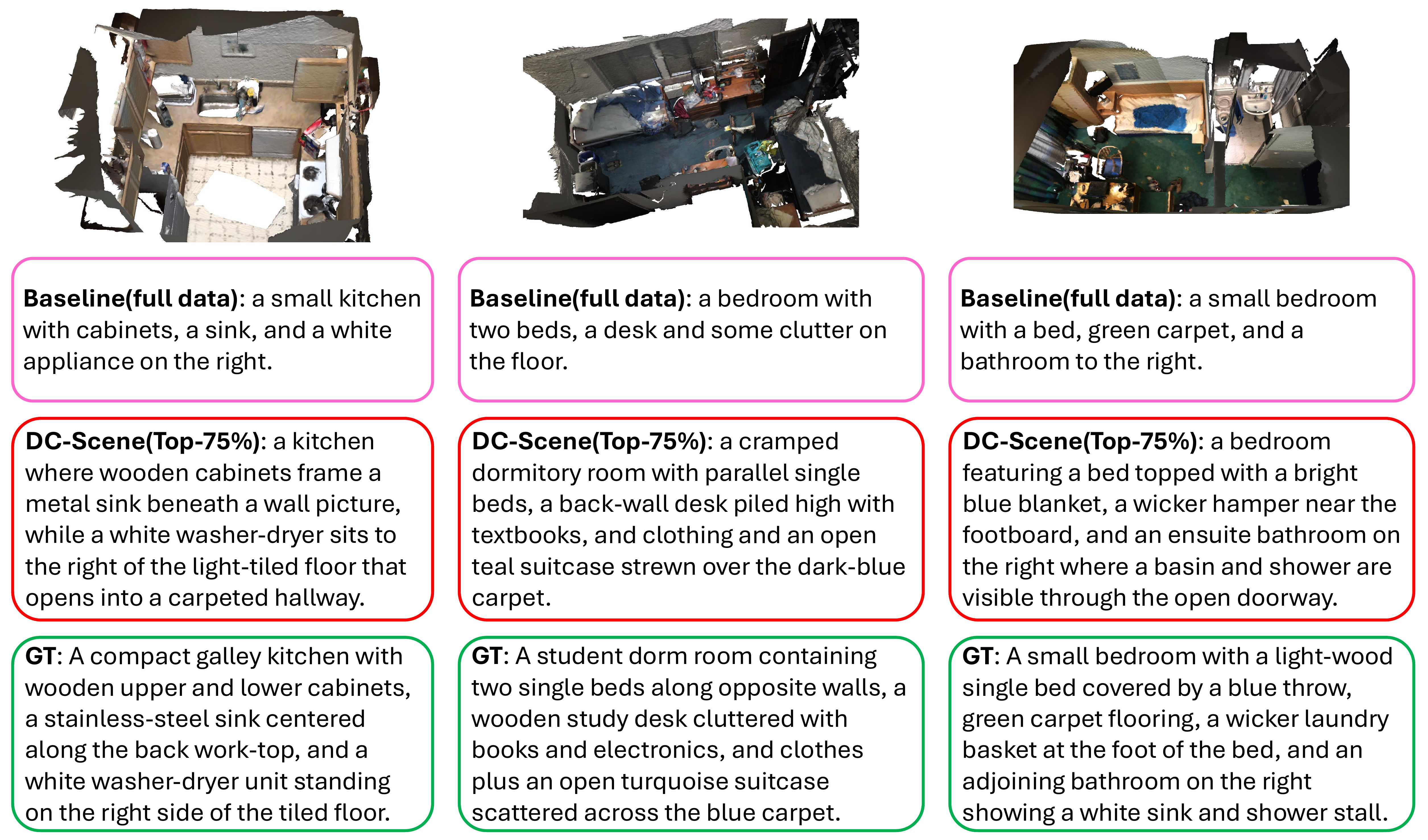}
    \caption{\textbf{Qualitative comparison of scene captions.} For three validation scenes from the ScanRefer~\cite{chen2020scanrefer} dataset, we present the rendered point cloud mesh (top row), followed by captions generated by three sources: the full-data baseline model (in pink), our \textbf{DC-Scene} model trained on the top-75\% DIQ samples (in red), and the human-annotated ground truth (in green).}
    \label{fig:visualization}
    \vspace{-0.4cm}
\end{figure}
As shown in Figure~\ref{fig:visualization}, we present three representative validation scenes from the ScanRefer~\cite{chen2020scanrefer} dataset to illustrate how the proposed \textbf{DC-Scene} curriculum enhances caption quality compared to a strong full-data baseline.
\section{Conclusion}
In this paper, we proposed \textbf{DC-Scene}, a data-centric learning framework designed to address the dual challenges of computational overhead and data scarcity in 3D scene captioning. Powered by a CLIP-based Dual-Indicator Quality (DIQ) filter and a three-stage curriculum scheduler, DC-Scene prioritizes semantically coherent training samples and introduces them in a pedagogically structured sequence. Specifically, our progressive curriculum learning strategy incrementally expands the training data from the top-25\% DIQ subset, through the top-50\% subset, ultimately reaching optimal performance at the top-75\% DIQ subset. This strategy significantly reduces training time while preserving, and even enhancing, performance. On ScanRefer and Nr3D, training with only the top-75\% DIQ subset achieves 86.10 CIDEr at IoU 0.25 and 53.60 CIDEr at IoU 0.50 using the 3D CoCa backbone—surpassing the full-data baseline with just one-third of the training epochs. Consistent improvements observed with the Vote2Cap-DETR++ backbone further demonstrate the generalizability and effectiveness of the proposed approach across different model architectures.
% \clearpage
% \bibliography{egbib}

\end{document}